\documentclass[10pt,twocolumn,letterpaper]{article}

\usepackage{cvpr}              

\definecolor{cvprblue}{rgb}{0.21,0.49,0.74}
\usepackage[pagebackref,breaklinks,colorlinks,allcolors=cvprblue]{hyperref}
\usepackage{graphicx}  
\usepackage{subcaption}
\usepackage{hyperref}
\usepackage{geometry}
\usepackage{lscape}
\usepackage{multirow}
\usepackage{threeparttable}
\usepackage{csquotes}

\def\paperID{*****} 
\def\confName{CVPR}
\def\confYear{2025}

\title{Diachronic Document Dataset for Semantic Layout Analysis}

\author{%
Thibault Clérice\textsuperscript{1}, Juliette Janès\textsuperscript{1}, Hugo Scheithauer\textsuperscript{1}, Sarah Bénière\textsuperscript{1}, \\%
Florian Cafiero\textsuperscript{2,3}, Laurent Romary\textsuperscript{1}, Simon Gabay\textsuperscript{4}, Benoit Sagot\textsuperscript{1} \vspace{.5em} \\ %
\textsuperscript{1} Inria, Paris, France, {\tt\small {firstname}.{lastname}@inria.fr} \\
\textsuperscript{2} Paris Sciences \& Lettres, Paris, France, {\tt\small florian.cafiero@chartes.psl.eu} \\
\textsuperscript{3} Center for Digital Humanities and Multilateralism,\\ Geneva Graduate Institute, Geneva, Switzerland \\
\textsuperscript{4} University of Geneva, Geneva, Switzerland, {\tt\small simon.gabay@unige.ch} 
}

\newcommand{\ladas}[1]{{\small\texttt{#1}}}
\newcommand{\smalltt}[1]{{\small\texttt{#1}}}
\newcommand{\datasetname}{the LADaS 2.0 Dataset}
\newcommand{\DatasetName}{The LADaS 2.0 Dataset}
\newcommand{\ladasname}{LADaS Dataset} 

\begin{document}
\maketitle

\begin{abstract}
We present a novel, open-access dataset designed for semantic layout analysis, built to support document recreation workflows through mapping with the Text Encoding Initiative (TEI) standard. This dataset includes 7,254 annotated pages spanning a large temporal range (1600-2024) of digitised and born-digital materials across diverse document types (magazines, papers from sciences and humanities, PhD theses, monographs, plays, administrative reports, etc.) sorted into modular subsets. By incorporating content from different periods and genres, it addresses varying layout complexities and historical changes in document structure. The modular design allows domain-specific configurations. We evaluate object detection models on this dataset, examining the impact of input size and subset-based training. Results show that a 1280-pixel input size for YOLO is optimal and that training on subsets generally benefits from incorporating them into a generic model rather than fine-tuning pre-trained weights. 
\end{abstract}

\section{Introduction}

\begin{figure}[t]
    \centering
    \includegraphics[width=\linewidth]{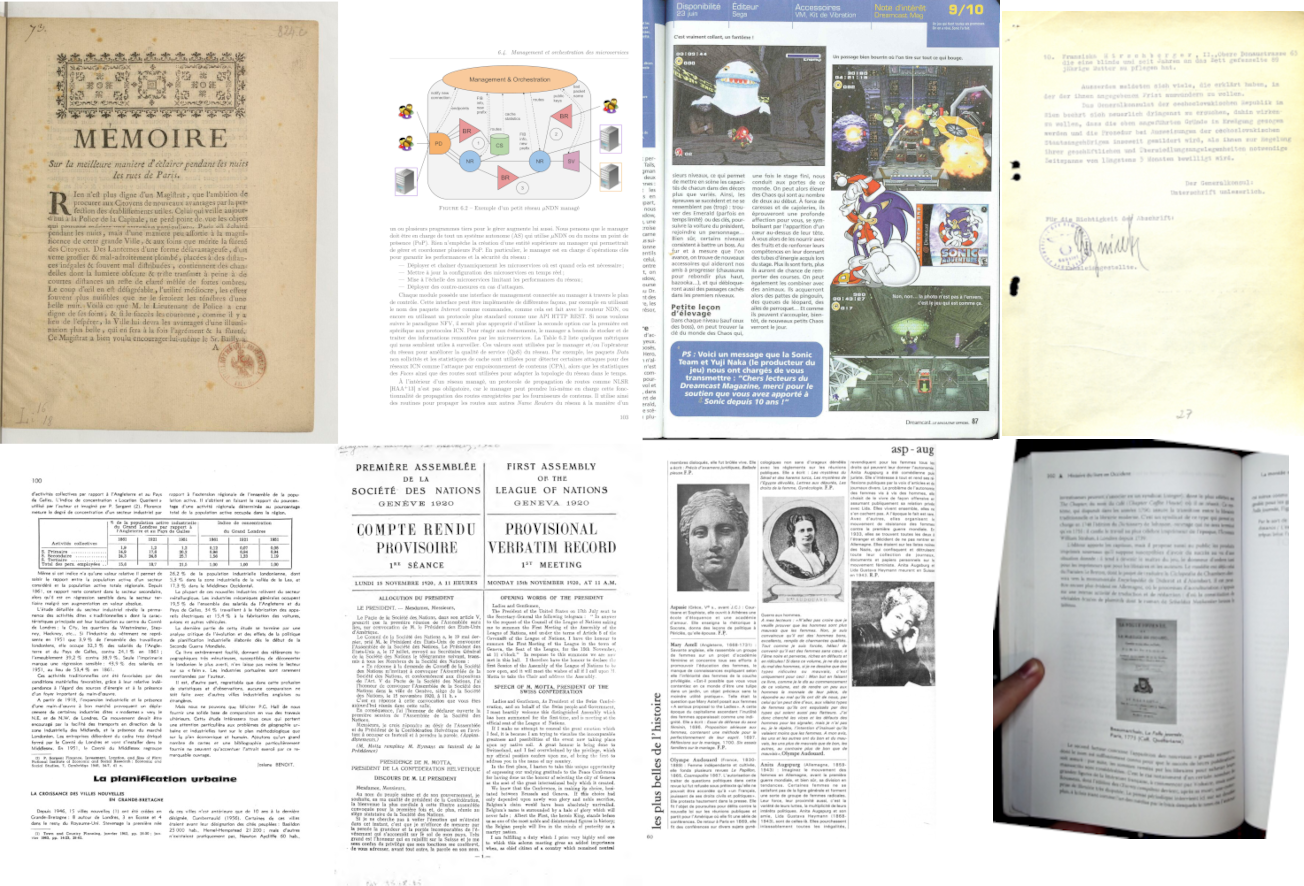}
    \caption{Examples of different layouts from subsets of the dataset.}
    \label{fig:images-subset}
\end{figure}

Optical Character Recognition (OCR) and Document Layout Analysis (DLA) are essential steps in converting analogue or born-digital documents into digital, machine-interpretable formats, particularly for documents where the text flow cannot be directly interpreted (e.g., PDFs). M$^{6}$Doc~\cite{cheng_m_2023} recently revisited the distinction between physical and logical layout analysis, where the former identifies document features (e.g., text, images) and the latter focuses on their semantic roles. While extensive datasets have been developed for both kinds of analysis (\cref{tab:datasets}), few address non-digital-born documents comprehensively. Among these, many focus heavily on highly repetitive layouts, such as those in Science, Technology, Engineering, and Mathematics (STEM) archival repositories exemplified by DocBank~\cite{li_docbank_2020}. In this context, object detection approaches and bounding box annotation methods have emerged as pivotal tools for DLA, aligning with document digitisation workflows while maintaining high processing efficiency.

Cultural Heritage (CH) institutions, along with Social Sciences and Humanities (SSH) repositories, and researchers working on historical documents, face persistent challenges in reconstructing diverse and diachronically varied materials. Existing DLA datasets rarely capture the temporal or structural diversity inherent to these fields. Digital repositories such as Gallica\footnote{\url{https://gallica.bnf.fr/}.} and Persée\footnote{\url{https://www.persee.fr/}.} often limit their outputs to page-level facsimiles, leaving researchers -- specifically in the Digital Humanities (DH) -- reliant on extensive post-processing and manual annotation to produce complete, structured documents. As documents are central to these researchers, semantic tagging of documents in these communities has been standardised through the Text Encoding Initiative (TEI) Guidelines~\cite{teiconsortium}, which offer XML schemas for encoding digital texts with rich semantic markup.

In this paper, we aim to bridge the gap between datasets built for DLA in computer vision (CV) conferences and DH practices by introducing \datasetname, an open, free, diverse, diachronic, document-reconstruction-centric dataset\footnote{\DatasetName{} will be provided on HuggingFace upon publication.}. Building upon the SegmOnto~\cite{gabay_segmonto_2021} controlled vocabulary and syntax for DLA -- which emphasised filtering primary text bodies from peripheral elements like running titles or marginal notes -- we extend its scope by generating a refined set of subclass types grounded in the TEI Guidelines. These subclasses are designed to facilitate document reconstruction while ensuring semantic consistency across diverse textual elements.

To rigorously evaluate and extend the applicability of SegmOnto’s taxonomy, we focus on documents from the 17th century onwards, encompassing a rich mixture of literary works, private and administrative reports, non-fiction texts, and catalogues (\cref{fig:images-subset} and \cref{tab:subsets}). Our dataset introduces 36 distinct classes, organised into 13 overarching types, designed to capture the multifaceted nature of documents in diachrony. Moreover, we maintain a metadata-rich version of the dataset, where each image is associated with detailed provenance information, a subset classification, as well as its publication date. These metadata allow users to reorganise and filter the dataset dynamically, aligning with various research objectives and expectations.

In this paper, we analyse existing DLA datasets and review object detection methods and benchmarks relevant to this task. We then present \datasetname, detailing its annotation guidelines, document selection process, production pipeline, and key statistics. Finally, we benchmark object detection models to evaluate their ability to capture our 36 classes, extending the evaluation to the comparison of performances on subsets versus a generic dataset to highlight the benefits of fine-tuning for specific use cases. 

The contributions of this paper are as follows:
\begin{itemize}
    \item a new set of classes for DLA that closely connect the former with document production in TEI;
    \item a new dataset of 7,254 documents for benchmarking models across various types of documents as well as different periods of time;
    \item a first benchmark and set of recommendations for providing models for such tasks based on the specificities of our dataset.
\end{itemize}

\section{Related Work}

\subsection{Datasets for Layout Recognition}

\begin{table*}
     \caption{Comparison of Layout Analysis Datasets. A.M stands for Annotation Method.}
    \scriptsize
    \centering
    \begin{threeparttable}
        \resizebox{\linewidth}{!}{
            \begin{tabular}{cllrrrll}
            	\toprule
            	Domain  & Dataset                                    & Documents Type                 & Digitisation & Pages  & Labels & A.M.      & Reusable \\  \midrule
            	\multirow{7}{*}{CV} & PrimA~\citep{antonacopoulos_realistic_2009} & Various Modern Documents       & Mixed        & 1240   & 10     & Mixed     & None     \\
            	                      & PubLayNet~\citep{zhong_publaynet_2019}      & Medical papers                 & No           & 360000 & 5      & Automatic & Yes      \\
            	                    & DocBank~\citep{li_docbank_2020}             & STEM papers                    & No           & 50000  & 12     & Automatic & None     \\
            	                    & Scibank~\citep{grijalva_scibank_2022}       & STEM papers                    & No           & 74435  & 12     & Automatic & Yes      \\
            	                    & DocLayNet~\citep{pfitzmann_doclaynet_2022}  & Various Modern Documents       & Mixed        & 80863  & 11     & Mixed     & Yes      \\
            	                    & M$^{6}$Doc~\citep{cheng_m_2023}             & Various Modern Documents       & Mixed        & 9080   & 74     & Manual    & Yes      \\
            	                    & ETD-ODV2~\citep{ahuja_new_2023}             & Thesis                         & Mixed        & 20 000 & 24     & Generated & None     \\ \midrule
            	\multirow{7}{*}{DH / CH} & SCUT CAB~\citep{cheng_scut-cab_2022}        & Chinese Ancient Books          & Yes          & 4000   & 27     & Manual    & Yes      \\
            	                    & American Stories~\citep{dell_american_2023} & Historic Press                 & Yes          & 2200   & 7      & Manual    & Yes      \\
            	                    & HJD~\citep{shen_large_2020}                 & 19th-20th Japanese Documents   & Yes          & 2271   & 7      & Manual    & Yes      \\
            	                    & Gallicorpora~\citep{pinche_between_2022}    & Literary Books and Manuscripts & Yes          & 981    & 15     & Manual    & Yes      \\
            	                    & HORAE~\citep{boillet_horae_2019}            & Books of Hours                 & Yes          & 500    & 13     & Manual    & Yes      \\
            	                    & Ajax~\citep{sven_page_2022}                 & 19th Critical Editions         & Yes          & 300    & 18     & Manual    & Yes      \\ \midrule
                                    & \DatasetName                               & Various Documents in Diachrony & Mixed        & 7,254  & 36     & Manual    & Yes      \\ \bottomrule
            \end{tabular}
        }
        \label{tab:datasets}
    \end{threeparttable}
\end{table*}

The PrimA~\citep{antonacopoulos_realistic_2009} dataset, introduced in 2009, was the first widely used document-focused layout analysis dataset and comprises 1,240 semi-automatically annotated images. Around 2020, large datasets, like PubLayNet~\citep{zhong_publaynet_2019} and DocBank~\citep{li_docbank_2020}, were introduced, focusing on scientific papers from ArXiv and PubMed, largely leveraging the possibilities of automatically annotating PDFs based on the LaTeX or XML available on these repositories. DocLayNet~\citep{pfitzmann_doclaynet_2022} and M$^{6}$Doc~\citep{cheng_m_2023} were developed shortly after. Their scopes were broadened to include a more diverse range of sources. Even though scientific articles still make up a large portion of document types, both datasets contain other kinds of modern documents such as financial reports, legal documents, magazines, or textbooks. The M$^{6}$Doc dataset is the first to include not only native PDF documents but also scanned documents and photographs of various kinds. Although English is still prevalent in most datasets, efforts are being made to include other scripts such as Japanese and Chinese (M$^{6}$Doc). Outside of traditional CV, and specifically in DH, most layout analysis datasets have been produced for specific projects focusing on a specific type of material: historical newspapers, monographs, manuscripts, and critical editions. Like M$^{6}$Doc and DocLayNet, time-consuming manual annotation campaigns are required to produce these datasets. These two parameters are why most datasets contain less than 1,000 images and at most 10,000. Only a few of the datasets have a broad reusable license (\cref{tab:datasets}).

The diverse sources in layout analysis datasets result in numerous different annotation guidelines with custom labels. Sometimes, a concise and common annotation system is used, describing a basic layout (\smalltt{text}, \smalltt{heading}, \smalltt{graphic}, \smalltt{list}, and \smalltt{table}) as for PubLayNet~\citep{zhong_publaynet_2019} or American Stories~\citep{dell_american_2023}. Other times, more specific labels are used, focusing on the semantic significance of each zone. For example, what might simply be labelled as \smalltt{text} in PubLayNet could be further categorised as a \smalltt{caption} or an \smalltt{abstract} in DocBank. The number of labels can even go up to 70, with the example of M$^{6}$Doc, which provides a detailed zone description with specific document labels or highly granular labels for titles (e.g. \smalltt{fourth-level title,} \smalltt{third-level question number}). In the DH community, the introduction of SegmOnto~\citep{gabay_segmonto_2021} in 2021 and its adoption led to a better ability to combine datasets: it was used and adapted by projects like Gallicorpora~\citep{pinche_between_2022} or Ajax~\citep{sven_page_2022}, enabling their interoperability.

The M$^{6}$Doc annotation system and the SegmOnto guidelines can be compared due to their similar labels and their shared characteristic of referring to a layout analysis guide. However, they differ in terms of depth and focus. SegmOnto focuses on distinguishing zones or bodies of text and as such provides general labels for various historical layouts with a three-level syntax such as \smalltt{Type:Subtype\#Numbering}, the type being a broad area of description whereas the subtype is used to specify, if needed, this area. SegmOnto only features the type level as a controlled vocabulary to denote the main body of text (e.g. \smalltt{MainZone}), specific margin elements (e.g. \smalltt{RunningTitleZone}, \smalltt{MarginTextZone}) or other specialised types of zones, such as for media, each with its own label (e.g. \smalltt{GraphicZone}, \smalltt{TableZone}). The different types can be described in more detail with specific project subtypes (\ladas{MainZone:Column}). In contrast, the M$^{6}$Doc guidelines delve deeper and offer more detailed descriptions with specific labels depending on the zone level and the semantics, for specific contemporaneous documents.

\subsection{Object Detection for Layout Recognition}

Of all object detection architectures, the most well-known are the YOLO ones which aim at achieving both high accuracy and high throughput. Its latest release, YOLOv11~\cite{khanam2024yolov11overviewkeyarchitectural}, has been released in October 2024.

This high throughput and accuracy added to the bounding-box compatible nature of most layouts have led, across fields ranging from DH to CV, to multiple benchmarks for various document layout analyses, including generic~\cite{yaltai}, domain-specific~\cite{najem2022page}, and partial~\cite{Chanda2020} applications. These models demonstrate superior adaptability to smaller datasets compared to R-CNN and mixed transformer approaches~\cite{kastanas2023documentaicomparativestudy}. As a result, studies have proposed adaptations of these architectures for document analysis, including YOLOv8~\cite{Deng2023} and, more recently, DocYOLO, based on YOLOv10~\cite{zhao2024doclayoutyoloenhancingdocumentlayout}. Both approaches emphasise domain-specific optimisations using contemporary datasets tailored for document layout evaluation. 

\section{\DatasetName}

\subsection{Annotations Guidelines}

While SegmOnto provides a set of zone types for distinguishing noise from the main body of text, it lacks clear specifications regarding the scope of annotation. For instance, it does not clarify whether the \smalltt{MainZone} should apply to an entire column or individual paragraphs. Additionally, it lacks tools for the normalised classification of sub-elements within the first level, such as paragraphs or lists. To address these gaps and construct our guidelines and class set, we selected subclasses (\cref{tab:classes}) based on the availability of a corresponding TEI XML class, the visual distinguishability of elements, and their relevance for post-processing information separation. We also simplify the syntax of SegmOnto through the use of \smalltt{-} to separate the first and second levels, instead of \smalltt{:} (Level1:Level2 is annotated Level1-Level2).

The most common type in our subset is the \ladas{MainZone}, which refers to the primary content-bearing element of a document page, as opposed to margins or illustrations for example. In \ladas{MainZone}, we distinguish the various elements that compose a text, including groups of lines (\ladas{MainZone:Lg}), paragraphs (\ladas{MainZone:P}), items in lists (\ladas{MainZone:Item}), character's speeches like in plays(\ladas{MainZone:Sp}), and headings (\ladas{PageTitleZone} for page-wide titles, \ladas{MainZone:Head} for titles embedded in text). 
Additionally, margin elements are described with specific labels, such as \ladas{NumberingZone}, \ladas{RunningTitleZone}, and \ladas{MarginTextZone}, with a particular focus on distinguishing between printed (\ladas{MarginTextZone:Notes}\footnote{\ladas{MarginTextZone:Notes} identify each note individually, to mirror its counterparts in \ladas{MainZone}.}) and manuscript notes at a second level (\ladas{MarginTextZone:ManuscriptAddendum}).

\begin{table}
\caption{\DatasetName{} created subtypes based on SegmOnto types, except for zones in Italics that are new to SegmOnto.}
\centering
\scriptsize
\begin{tabular}{lp{.5\linewidth}}
	\hline
	Zones Type                 & \ladasname Subtypes                                        \\ \hline
    \textit{FormZone}          & \\
	MainZone                   & Head, P, Lg, Sp, List, Entry, Date, Signature, Maths, Other \\
	MarginTextZone             & Notes, ManuscriptAddendum                                  \\
	\textit{FigureZone}        & Head, Figdesc                                              \\
	GraphicZone                & Head, Figdesc, TextualContent, Part, Decoration     \\
	TableZone                  & Head                                                       \\
	PageTitleZone              & Index                                                      \\
	StampZone                  & Sticker                                                    \\
\end{tabular}
\label{tab:classes}
\end{table}

We applied the same level of granularity to the zones for tables, graphics, and figures (\ladas{TableZone}, \ladas{GraphicZone}, and \ladas{FigureZone}). Sublevels like \ladas{Head}, \ladas{FigDesc}, \ladas{TextualContent}, or \ladas{Part} provide crucial context for tabular and graphical elements, thus allowing to represent semantic textual hierarchy within graphical elements. The specific subtype \ladas{Decoration} provides a semantic description for ornamental \ladas{GraphicZone}, distinguishing it from other \ladas{GraphicZone}s that contain content. This approach ensures a nuanced representation of the relationships between textual content and graphical components, contributing to a more comprehensive understanding of document layout and structure in our dataset.

When applying a layout analysis vocabulary to modern documents, such as videogames magazines or theses, we encountered specific layouts that had not been previously considered by SegmOnto. As a result, we introduced a few additional first-level labels: \ladas{FigureZone} for programming excerpts, \ladas{FormZone} for form in magazines, and \ladas{AdvertisementZone} for advertisements. These elements, which are specific to modern and contemporaneous documents, represent a combination of typographical and graphical features unique to them.

Since our additions to SegmOnto aim to reconstruct the flow of a complete document, we address challenges posed by run-on paragraphs or elements, as well as text disrupted by intervening figures. After careful deliberation, we decided to treat any run-on element belonging to the same category within its SegmOnto type (e.g., \ladas{MainZone} and \ladas{MarginTextZone}) by creating a \ladas{Continued} subtype. Using a single subtype per zone type acknowledges the difficulty of distinguishing, without the context of the preceding page, between a run-on paragraph and a run-on catalogue entry or list item. By applying reading order heuristics -- whether navigating between columns or transitioning across pages during document reconstruction -- we effectively merge any \ladas{[...]Continued} with its preceding element of the corresponding type, ensuring coherence in the final document structure.

\subsection{Annotation Campaign}

\paragraph{Content Selection and Diversity}

\begin{table*}[t]
	\caption{\DatasetName's subsets. F/NF stands for Fiction/Non-fiction, A/NA for Academic/Non-Academic. RH for Random Harvesting, List means harvesting was based on metadata, \enquote{- 1} or \enquote{- 2} indicates the maximum number of pages per single document. Numbers for splits are given in number of pages.}
    \centering
    \scriptsize
		\begin{tabular}{lllcccl|rrrr}
Subset         & Provenance    & Acquisition & Status    & Fiction & Academic & Century & Train & Valid & Test & Total \\ \midrule
Admin. Rep,    & Various       & List - 3    & Mixed     & NF    & NA    & 19-21 & 100   & 30    & 99   & 229   \\
Catalogues     & INHA-BNF      & Donation    & Digitised & NF   & NA    & 19-20 & 1072  & 265   & 100  & 1437  \\
Fingers        &               & Donation    & Digitised & Mix. & NA    & 21    & 51    & 6     & 43    & 100   \\
Magazines Tech &               & List - 1    & Digitised & NF   & NA    & 20-21 & 194   & 32    & 104   & 330   \\
Monographies   & Gallica       & List - 1    & Digitised & Mix. & NA    & 17-20 & 1689  & 203   & 100   & 1992  \\
Others         &               & Production  & Digitised & NF   & NA    & 21    & 5     & 1     & 0     & 6     \\
Persée         & Persée        & RH - 1      & Digitised & NF   & A     & 20    & 985   & 128   & 117   & 1230  \\
Picard         & ARLP          & Donation    & Mixed     & F    & NA    & 21    & 87    & 6     & 4     & 97    \\
Romans 19      & Gallica       & List - 1    & Digitised & F    & NA    & 19    & 103   & 36    & 101   & 240   \\
Théâtre        & Gallica       & List - 1    & Digitised & F    & NA    & 17-20 & 540   & 104   & 106   & 750   \\
Thèses         & Theses.fr     & RH - 2      & Dig. Born & NF   & A     & 21    & 536   & 92    & 117   & 745   \\
Typewriter     & DAHN-EHRI     & Donation    & Digitised & NF   & NA    & 20    & 81    & 9     & 8     & 98    \\ \midrule
All            &               &             &           &      &       &       & 5443  & 912   & 899  & 7254  \\ \bottomrule
		\end{tabular}
	\label{tab:subsets}
\end{table*}
\DatasetName{} contains a total of 7,254 document images from various periods, ranging from the 17th century to the present day, categorised into ten subsets based on their content and provenance (\cref{tab:subsets}). The dataset includes a wide variety of content, from fiction -- such as prose, poetry, and drama (Monographies, Picard, Théâtre subsets) -- to non-fiction, including administrative documents (Typewriter and Administrative Reports subset), academic papers (Persée and Thèses subset), magazines about new technologies and video games, and 19th-century catalogues of numismatics or art galleries. While the majority of the dataset is in French, there is a small presence of other languages (Picard, English, Latin, etc.) and scripts (e.g., Japanese, Arabic), specifically in the academic documents. Most of the documents published after 2000 are digital-born, extracted from PDFs, while the others are printed, and one subset consists of typewritten documents.

Three methods of acquisition were used to compile \datasetname: data donations from partners (e.g., the Picard subset from the Agence Régionale de la Langue Picarde (ARLP),\footnote{\url{https://languepicarde.fr/}} or the Catalogues subset from the French National Institute for Art History (INHA), and the French National Library, (BnF)), targeted randomised harvesting of portals using pre-filtered lists of documents (such as the Gallica subsets), and randomised harvesting (RH) from repositories.

Finally, the Fingers subset was created to introduce noise into the dataset. We deliberately digitised books using book scanners or phone cameras in a suboptimal manner, leaving fingers, background clutter, and bent pages visible in the camera shot. This subset is designed to replicate common on-site issues and allow models to address diverse needs, such as helping researchers or students extract text from books in a library setting.

\paragraph{Annotation Process}

The data was annotated by the authors of this paper, along with additional trained annotators who were hired and instructed according to the annotation guidelines. The annotation process was conducted on the Roboflow platform~\cite{dwyer2022roboflow}, with initial pre-annotations generated by models trained on prior versions of the dataset after the first few hundred pages.\footnote{Around twenty models have been trained for pre-annotation in the course of a year.} Each annotation underwent review by a second annotator, and any disagreements prompted discussion among the guideline authors to ensure consistency.

\subsection{Statistics}

\begin{figure}
    \centering
    \includegraphics[width=\linewidth]{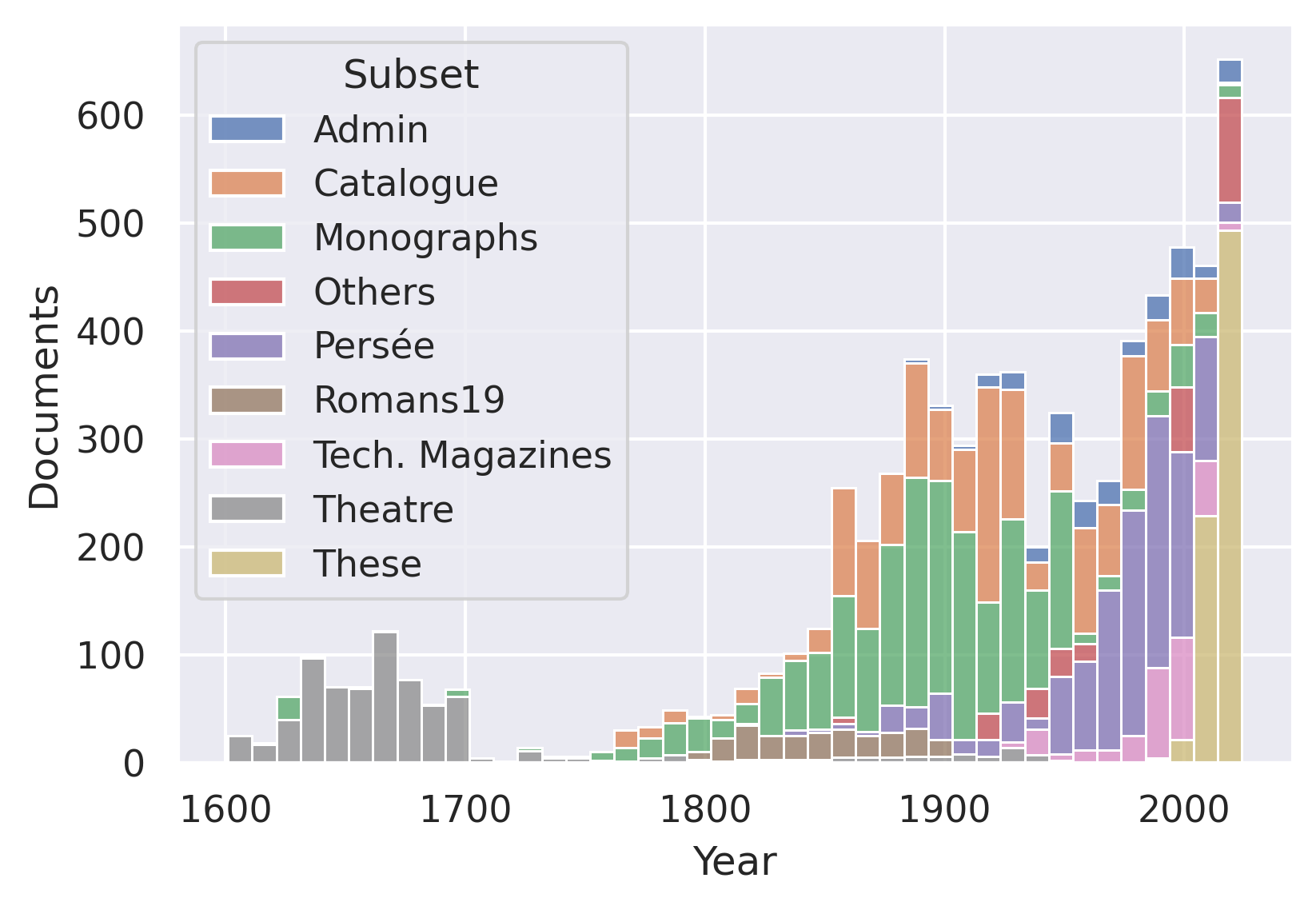}
    \caption{Distribution over time of documents based on our subsets.}
    \label{fig:time-documents}
\end{figure}

\begin{figure}
    \centering
    \includegraphics[width=\linewidth]{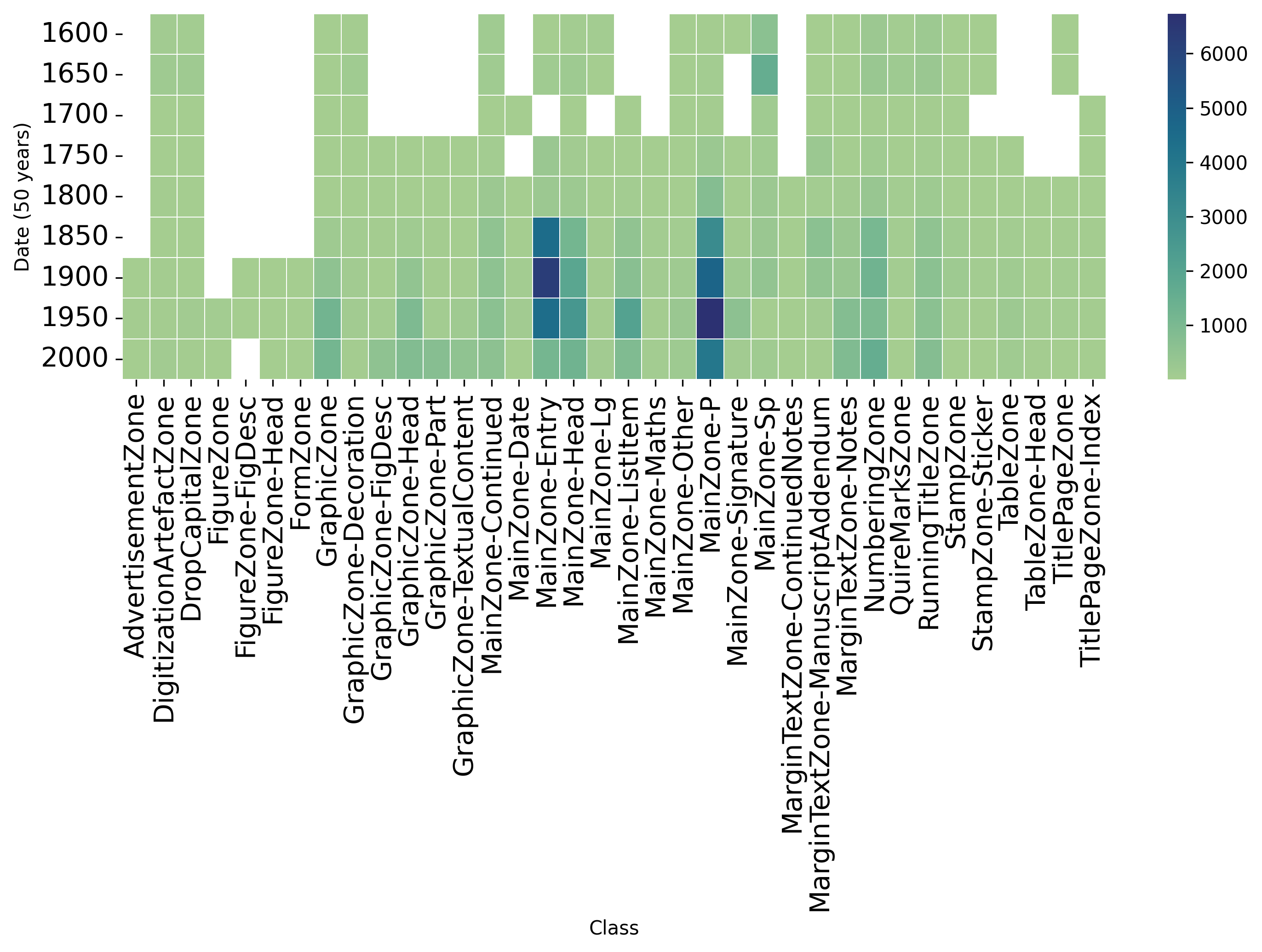}
    \caption{Distribution of class instances over time.}
    \label{fig:time-classes}
\end{figure}

The resulting corpus comprises 7,254 pages and 81,766 instances, showing a distinct peak at the turn of the 19th century and another around the 2000s. The latter is largely attributed to the PhD theses subset, whose collection predominantly utilised digital repositories. The most comparable dataset in terms of semantic labelling approach and page count, M$^{6}$Doc, contains 9,080 images and 237,116 objects. The difference in the instance-to-page ratio may stem from the granularity of M$^{6}$Doc's approach (e.g., its inclusion of classes such as ``weather report,'' which in our case would be simplified as \ladas{GraphicZone}) or the nature of its corpus, which incorporates more magazines (1,000 documents) and newspapers (500 documents). These document types typically exhibit the richest and most complex layouts, as observed in our experience.

Our subsets are unevenly distributed; however, six subsets include test sets with approximately 100 test pages each (Admin. Rep., Catalogues, Tech. Magazines, Monographies, Persée, Romans-19, Théâtre, Thèses). Among these, three subsets feature training sets comprising around 1,000 pages or more (Catalogues, Monographies, Persée).

Subsets sourced from donations, as well as PhD theses and elements available only in recent times (e.g., Tech. Magazines), exhibit the most skewed distributions in terms of temporal coverage (e.g., Théâtre specifically focuses on the 17\textsuperscript{th} century, while Picard primarily spans the 21\textsuperscript{st} century; \cref{fig:time-documents}). Nevertheless, in terms of class distribution (\cref{fig:time-classes}), the most common classes are consistently represented from at least 1750 onwards (\ladas{MainZone-Entry}, \ladas{-Date}, \ladas{-Lg}, etc.). Certain classes, such as \ladas{MainZone-P}, are present across the entire dataset, while more marginal zones, including \ladas{Numbering}, \ladas{QuireMarks}, and \ladas{RunningTitle}, are also well-documented.

\section{Experiments with Object Detection}

\subsection{Generic models}
\label{genmod}

Building on the extensive experimental results of M$^{6}$Doc \cite{cheng_m_2023}, we focus on the impact of input image size rather than comparing different models. We train three series of YOLOv11 models \cite{yolo11_ultralytics} with images resized to maximum dimensions of 640, 960, and 1280 pixels, across five model sizes (\textbf{n}ano, \textbf{s}mall, \textbf{m}edium, \textbf{l}arge, and e\textbf{x}tra-large). Our hypothesis is that certain boundaries or classes are harder to detect in low-resolution images, a challenge raised by annotators struggling to distinguish marginal notes in 640-pixel scans. For this experiment, Théâtre, Admin. Rep., and 19th-century Novels are excluded from the training pool to enable a second set of experiments in \cref{domspec}. We evaluate our best model against the DocYOLO~\cite{zhao2024doclayoutyoloenhancingdocumentlayout} architecture, published after M$^{6}$Doc.

\paragraph{Set-up}

The models are trained with the \smalltt{Ultralytics} library (8.3.8), on a single GPU (RTX8000) for all models except for the 1280-pixel extra-large (x) model which required two. The batch size (16), number of epochs (100), seed (42), augmentations (mostly rotations, contrast, and sheer), lr (0.01), and other parameters are the same across the experiments. All other parameters are the default from \smalltt{ultralytics}. DocYOLO was trained from the pre-trained weights with the same parameters except for the learning rate (0.02 yielding better results) on 2 GPUs.

\paragraph{Results}

\begin{figure}[t]
    \centering
    \includegraphics[width=1\linewidth]{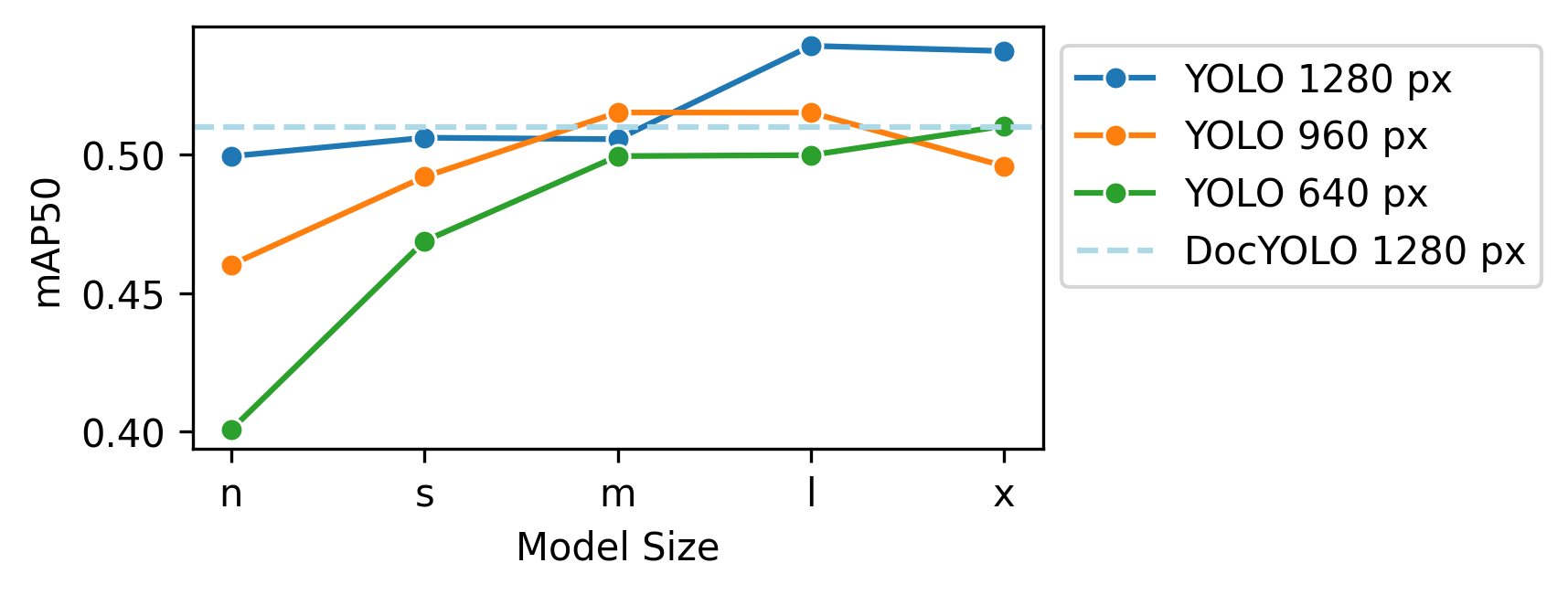}
    \caption{Curve of the mAP50 based on the input size and the model size.}
    \label{fig:yolo-configurations}
\end{figure}

\begin{figure}[t]
    \centering
    \includegraphics[width=1\linewidth]{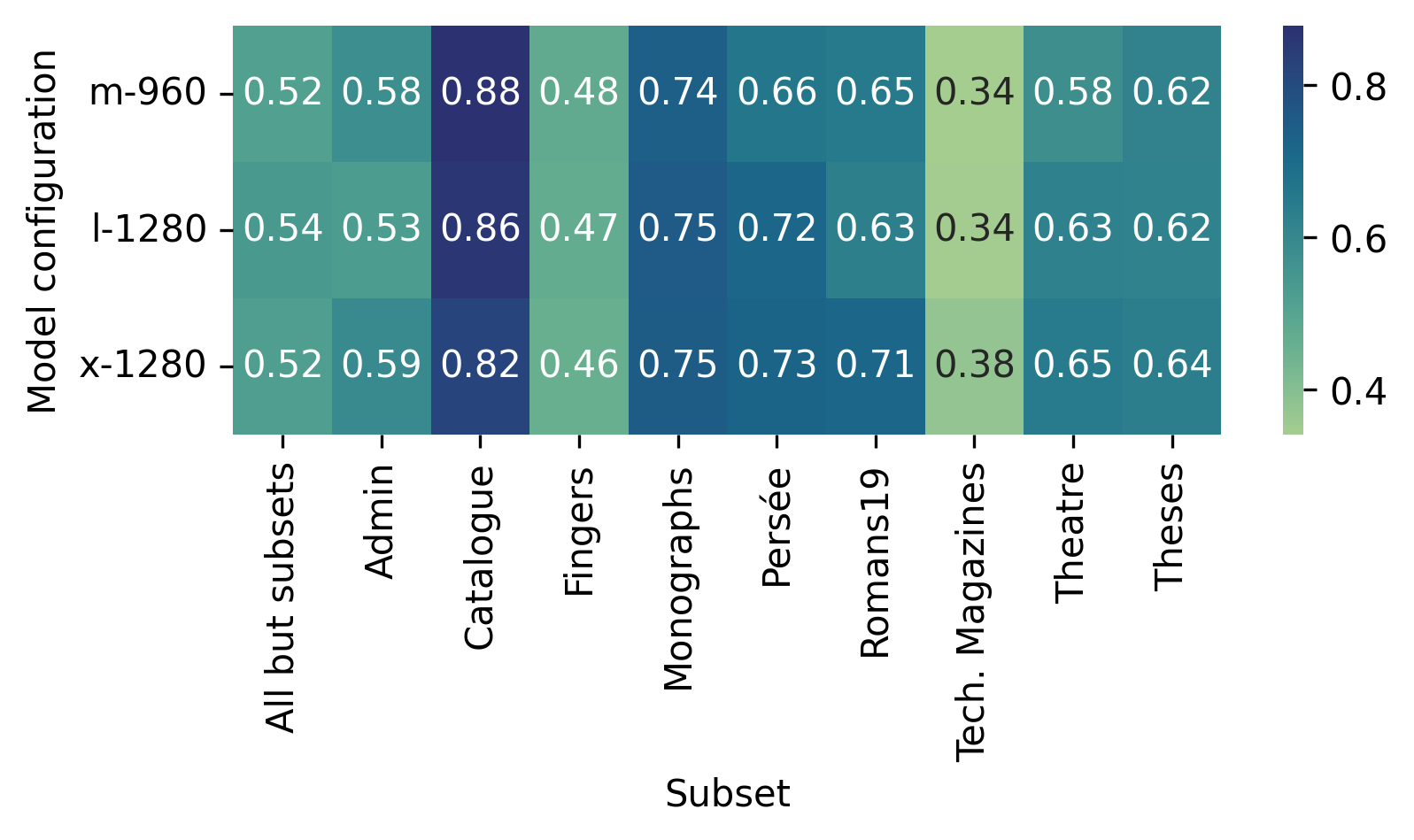}
    \caption{mAP50 across YOLOv11 best configurations starting from the medium model on each 100 pages subsets.}
    \label{fig:subset-results}
\end{figure}

As expected, increasing the input size generally improves the macro averages of mean average precision (mAP) scores (\cref{fig:yolo-configurations}). Models with a 1280-pixel input consistently outperform those with 960-pixel inputs, which in turn surpass the 640 ones. However, performance gains plateau with larger models, as the extra-large models provide minimal improvements in mAP50 and even show declines at 960 and 1280-pixel input sizes.\footnote{Additional tests with different seeds were conducted to confirm the consistency of this phenomenon.} The large model with a 1280-pixel input achieves the best results across all configurations, while DocYOLO underperforms despite its layout-specific enhancements to YOLOv10. 

Two subsets -- Fingers and Tech. Magazines -- substantially lower overall scores (\cref{fig:subset-results}). Performance in the Fingers subset is likely affected by noise and curvature, whereas the Tech. Magazine subset suffers from high visual complexity and limited training data. New augmentation strategies, such as merging plain white or black backgrounds and incorporating fingers, could help mitigate these challenges.

\subsection{Domain-specific models}
\label{domspec}

While our dataset is designed as a mix of diverse domains, it also includes rich metadata -- such as publication time (available for 95\% of the documents), domain, and data provider -- enabling users to focus on specific subdomains. This allows training models tailored to these subsets. We analyse three distinct subsets with varying relationships to the main dataset: Théâtre is dominated by \ladas{MainZone:SP}, which is largely absent in other subsets; Admin. Rep blends digital-born and cultural heritage content, featuring layouts typical of reports with familiar structures like headings and paragraphs; Romans-19 is closest to the main dataset, sharing many features with the Monographies subset and displaying a limited but common set of features.

\paragraph{Set-Up}

Using the same hyperparameters as \cref{genmod} and the same environment, we focus on the best-performing model size (L) and input size (1280) to train three models per subset. The first model combines the training set from \cref{genmod} with the subset data, assessing the efficiency of merging subsets for a generic model. Additionally, we fine-tune two models using only the subset data: one starting from the raw YOLOv11-L weights from Ultralytics and another from the previously trained Exp-1 model. As a baseline, we include the score from the initial experiment using the same parameters without exposure to the subset data.

\paragraph{Results} Based on the mAP50 values, the results presented in \cref{tab:exp2-results} indicate a clear advantage for training generic models across the majority of subsets, despite potentially significant differences in individual class characteristics or training samples, particularly within the Théâtre subset. The Admin. Rep. subset could be seen as an exception but the fine-tuned Exp-1-L model only achieves marginally higher performance -- by less than one percentage point -- compared to the generic model. Finally, our baseline actually beats a fine-tuned model on YOLOv11-L in the context of Romans-19, as the general layout of such documents is shared across multiple subsets, including Monographies.

\begin{table}[t]
\caption{Fine-tuning experiments on domain-specific subsets.}
\resizebox{\linewidth}{!}{
\begin{tabular}{lllrrrrrr}
\toprule
Subset     & Base Model     & Dataset    & Precision      & Recall          & mAP50          \\ \midrule
Théâtre    & Baseline       & -          & 0.578          & 0.633           & 0.626          \\ 
Théâtre    & YOLOv11-L       & All        & \textbf{0.758} & \textbf{0.766}  & \textbf{0.779} \\ 
Théâtre    & Exp-1-L        & Subset     & 0.681          & 0.68            & 0.687          \\ 
Théâtre    & YOLOv11-L       & Subset     & 0.662          & 0.574           & 0.642          \\ \midrule
Romans-19  & Baseline       & -          & 0.607          & 0.591           & 0.631          \\ 
Romans-19  & YOLOv11-L       & All        & 0.763          & \textbf{0.633}  & \textbf{0.735} \\ 
Romans-19  & Exp-1-L        & Subset     & \textbf{0.85}  & 0.585           & 0.667          \\ 
Romans-19  & YOLOv11-L       & Subset     & 0.429          & 0.484           & 0.486          \\ \midrule
Admin. Rep.     & Baseline       & -          & 0.576          & 0.582           & 0.529          \\ 
Admin. Rep.     & YOLOv11-L       & All        & 0.666          & 0.658           & 0.695          \\ 
Admin. Rep.     & Exp-1-L        & Subset     & 0.641          & \textbf{0.7}    & \textbf{0.703} \\ 
Admin. Rep.     & YOLOv11-L       & Subset     & \textbf{0.714} & 0.526           & 0.571          \\ \bottomrule
\end{tabular}}
\label{tab:exp2-results}
\end{table}

\section{Conclusion}

In this paper, we presented \datasetname, a new free and open dataset designed to bridge the gap between computer vision and digital humanities. It also offers the first diachronic dataset with extensive semantic annotations for document layout analysis. Its modular structure and date metadata enable flexible training splits and model customisation. We provided comprehensive benchmarks for both general-purpose annotations and subset-specific analyses, demonstrating the dataset's versatility and adaptability to various tasks.

Future work includes expanding temporal coverage with non-digital-born PhDs, diverse publications, and 17th-19th century literature. Additionally, we plan to explore artificial image blending techniques to improve on-site DLA efficiency, potentially reducing the need for manual annotation for the Fingers subset.

{
    \small
    \bibliographystyle{ieeenat_fullname}
    \bibliography{main}
}


\end{document}